\documentclass{article}
\usepackage{spconf,amsmath,graphicx}

\usepackage{times}
\usepackage{latexsym}
\usepackage{multirow}
\usepackage{graphicx}
\usepackage{enumitem}
\usepackage[linesnumbered,ruled,vlined]{algorithm2e}
\usepackage{amsmath}
\usepackage{amssymb}
\usepackage{latexsym}
\usepackage[inkscapelatex=false]{svg}
\usepackage{url}

\SetKwInput{KwInput}{Input}                
\SetKwInput{KwOutput}{Output}              

\usepackage{amsthm,amsmath,amssymb}
\usepackage{mathrsfs}

\usepackage{makecell}
\usepackage{color}
\usepackage{booktabs}
\usepackage{hyperref}
\usepackage{float}
\usepackage[all]{nowidow}
\usepackage[T1]{fontenc}
\usepackage{microtype}

\title{Terminology-aware Medical Dialogue Generation}
\name{Chen Tang\textsuperscript{1,2*}, Hongbo Zhang\textsuperscript{2*}, Tyler Loakman\textsuperscript{2}, Chenghua Lin\textsuperscript{2$\dagger$}
\footnotemark[1],
Frank Guerin\textsuperscript{1}}
\address{\textsuperscript{1}Department of Computer Science, The University of Surrey, UK \\
\textsuperscript{2}Department of Computer Science, The University of Sheffield, UK \\
\texttt{\{chen.tang,f.guerin\}@surrey.ac.uk} \\
\texttt{\{hzhang183,tcloakman1,c.lin\}@sheffield.ac.uk}
}

\begin{document}
%
\maketitle

\renewcommand{\thefootnote}{\fnsymbol{footnote}} 
\footnotetext[1]{Equal contribution.} 
\footnotetext[2]{Corresponding author.} 

%
\begin{abstract}
Medical dialogue generation aims to generate responses according to a history of dialogue turns between doctors and patients. Unlike open-domain dialogue generation, this requires background knowledge specific to the medical domain. Existing generative frameworks for medical dialogue generation fall short of incorporating domain-specific knowledge, especially with regard to medical terminology. In this paper, we propose a novel framework to improve medical dialogue generation by considering features centered on domain-specific terminology. We leverage an attention mechanism to incorporate terminologically centred features, and fill in the semantic gap between medical background knowledge and common utterances by enforcing language models to learn terminology representations with an auxiliary terminology recognition task. Experimental results demonstrate the effectiveness of our approach, in which our proposed framework outperforms SOTA language models. Additionally, we provide a new dataset with medical terminology annotations to support research on medical dialogue generation. Our dataset and code are available at \url{https://github.com/tangg555/meddialog}.
\end{abstract}
%

\section{Introduction}
\label{sec:intro}

The goal of telemedicine is to provide patients with digital access to medical information, particularly as a first port-of-call where access to a medical professional may be limited. 
In order to better handle the growing demand for quick and easy access to healthcare services~\cite{zeng-etal-2020-meddialog,zhou-etal-2021-generation,abd2017analysing}, there is increasing research on Medical Dialogue Generation, which aims to assist telemedicine by automatically generating informative responses given a history of dialogues between the patient and doctor as input \cite{wootton2017introduction}.  
Such consultation dialogues typically contain a great amount of domain-specific knowledge, such as that relating to diseases, symptoms, and treatments. Therefore, without background knowledge of relevant medical terminologies, conventional language models tend to struggle to understand the semantics of medical dialogues and generate medically relevant responses~\cite{tang2022recent}. 
 
\begin{figure}[tb]
\centering
\includegraphics[width=0.8\columnwidth]{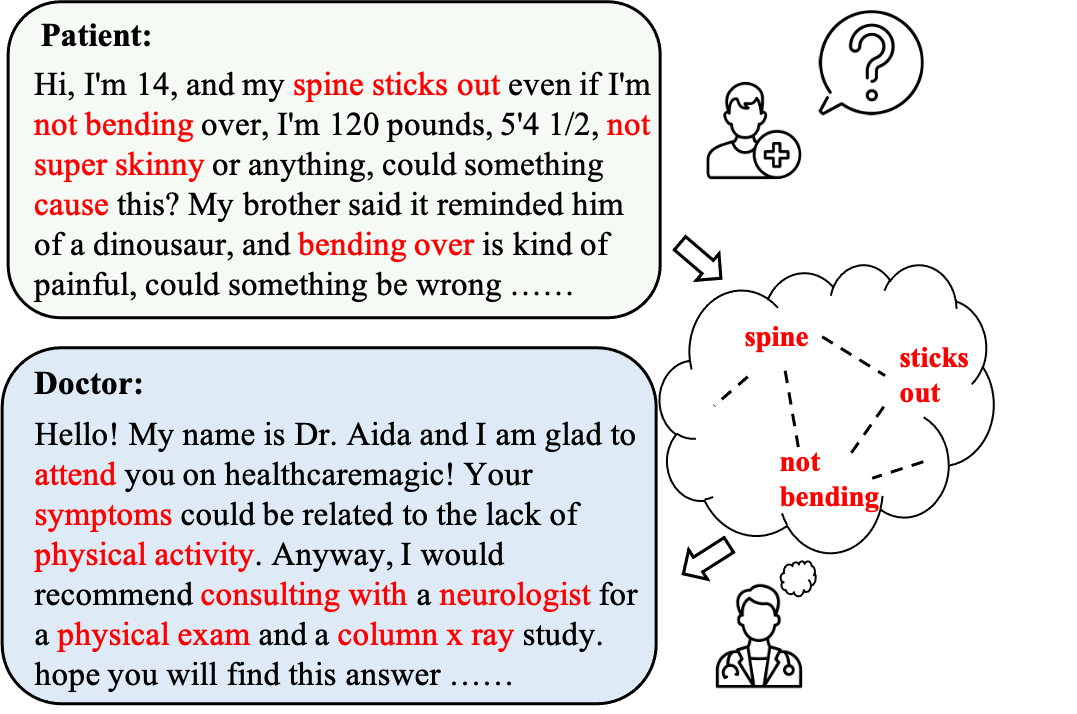}
\caption{A medical dialogue example from the provided dataset. The phrases in red denote terminology-related expressions that have been automatically annotated by our framework, which include Symptoms (sticks out), Examines (physical exam), Diseased parts (spine), and others.}
\label{fig: example}
\end{figure}

From an application perspective,  it is crucial for telemedicine dialogue systems to understand the medical issues of a patient, and provide informative suggestions as a ``doctor''. As illustrated in Figure~\ref{fig: example}, the dialogues between doctors and patients usually consist of both medically relevant and irrelevant expressions, 
e.g. ``\textit{My brother said it reminded him ...}''. The medically irrelevant expressions, to some extent, can be considered as noise, as they are prevalent in the dialogue
and may misguide a language model in understanding the medical context, consequently biasing the model towards generating responses that are unrelated to the medical issue of interest. Therefore, it is important to inform the language model of the most important aspects of the context, which is the medical terminology  (e.g. \textit{``spine sticks out'', ``not bending'', ``not super skinny''} in Figure~\ref{fig: example}), so that the model can learn the distribution of important medical features. 

To tackle medical domain-specific generation, prior works mainly explore two technical paradigms: (1) Implement pre-training on large-scale medical corpora, and then fine-tune language models on the target domain so that the language models can improve their performance on medical dialogues \cite{zeng-etal-2020-meddialog, zhou-etal-2021-generation}; and (2) Inject medical knowledge into language models with additional training targets to improve language understanding \cite{wei2018task,meddg,wei-etal-2018-task,lin2019enhancing,xu2019end}. 
The second paradigm has received increasing attention due to the effectiveness of knowledge injection. However, existing works all employ traditional language models (e.g. LSTMs) as the main network architecture with limited capacity. 
To our knowledge, there is currently no framework that employs and empowers large pre-trained models to incorporate medical terminology knowledge to improve medical dialogue generation.  

In order to improve the medical terminology understanding of language models, we propose a novel framework that fills in the medical knowledge gap of conventional encoders with additional terminology-aware training. It trains the neural networks to learn the feature distribution by enforcing the neural encoder to understand the medical terminologies from the input. Due to the lack of available large-scale medical dialogue corpora with annotated terminology, we develop an automatic terminology annotation framework, as well as a corresponding dataset for the purpose of further study (see our repository). The experimental results show superior performance on a range of metrics compared to SOTA language models, demonstrating the effectiveness of our proposed framework and the importance of medical terminology in medical dialogue generation.

Our contributions can be summarised as follows: (I) A novel pre-trained language model-based framework is proposed for terminology-aware medical dialogue generation; (II) An automatic terminology annotation framework and large-scale medical dialogue corpus with annotated terminology is provided; and (III) Extensive experiments demonstrate our framework achieves a substantial improvement over SOTA language models, owing to the better semantic understanding contributed by the enhancement of terminological knowledge.

\vspace{-1mm}
\section{Methodology}
\vspace{-1mm}
\begin{figure*}[tb]
\centering
\includegraphics[width=0.80\linewidth]{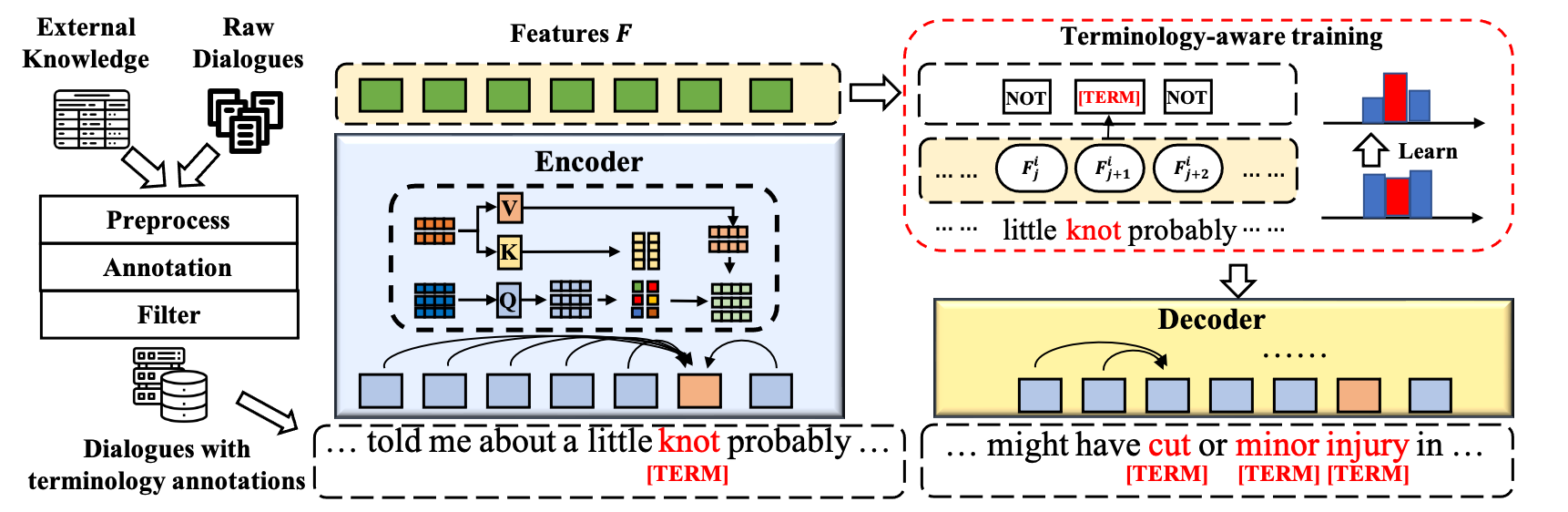}
\caption{The overview of our proposed framework. We choose BART as the base language model.}
\label{fig:overview}
\end{figure*}


Our proposed framework is illustrated in \autoref{fig:overview}, which includes an automatic terminology annotation component and a terminology-aware training component.

\vspace{-1mm}
\subsection{Task Definition}
\vspace{-1mm}
\label{sec:task}
We formulate our task as follows: the given inputs are in the form of a text sequence $ X = \{x_1, x_2, ..., x_n\}$, which consists of a patient's question alongside the collection of prior dialogue turns between the doctor and the patient. The goal of the task is to generate a response $ Y = \{y_1, y_2, ..., y_m\}$  (as a doctor) by modeling the conditional probability distribution $ P(Y|X)$.

\vspace{-1mm}
\subsection{Terminological Knowledge Enhancement}
\vspace{-1mm}

\noindent\textbf{Terminology Representations.}~~We employ an automatic annotation framework to annotate terminology-related tokens contained in dialogues (of both $X$ and $Y$) via inserting a special token $\mathrm{[TERM]}$. Take $X$ as an example:
\begin{equation}
    X_{\gamma} = \text{Flatten}(\text{Identify}(X))
\end{equation}
\begin{align}
\begin{split}
{x_i} = \left \{
\begin{array}{ll}
    {x_T, x_i},           & x_i\text{ is term} \\ 
    {x_i},               & \text{otherwise}
\end{array}
\right.
\end{split}
\end{align}
where $x_T$ denotes $\mathrm{[TERM]}$ and $X_{\gamma}$ denotes the input sequence with additional terminology annotations. \textit{Identify} denotes a distant supervision function for identifying the terminology tokens contained in plain text. It extracts all tokens of the input data that match the terminology word list\footnote{\url{https://github.com/glutanimate/wordlist-medicalterms-en}}, and then derives terminology phrases by assuming that adjacent tokens hold some form of relationship and constitute a terminological phrase. This approach largely eases the annotation process and avoids noise introduced by errors of normalisation. The annotated input is then flattened to be a text sequence for encoding. E.g., ``\textit{there is infection on hand}'' will be processed as ``\textit{there is} $\mathrm{[TERM]}  $\textit{infection on hand}''.


\begin{table}[tb]
\scriptsize \centering
\resizebox{0.8\linewidth}{!}{
\begin{tabular}{l|ccc}
\toprule
\textbf{Datasets}&\textbf{Train}&\textbf{Val}&\textbf{Test}\\
\midrule
\textbf{\# Dialogues} & 221,600 & 12,351 & 12,354 \\
\textbf{\# Words} & 36,700,836 & 2,041,364  & 2,052,748  \\
\textbf{\# Terms (words)} & 10,240,061 & 569,155 & 572,286 \\
\midrule
\textbf{Avg. \# Words in Input Text} & 104.15 & 103.84 & 103.84 \\
\textbf{Avg. \# Utterances in Input Text} & 1.19 & 1.20 & 1.19  \\
\textbf{Avg. \# Terms in Input Text} & 22.04 & 21.98 & 21.98 \\
\midrule
\textbf{Avg. \# Words in Output Text}  & 105.85 & 104.31 & 105.92 \\
\textbf{Avg. \# Utterances in Output Text} & 1.55 & 1.23 & 1.56 \\
\textbf{Avg. \# Terms in Output Text} & 24.05 & 22.24 & 24.10 \\
\bottomrule
\end{tabular}
}
\caption{Data statistics of the Medical English Dialogue Corpus with annotated medical terminologies (abbr. Terms).}
\label{tab:data_stat}
\end{table}

\noindent\textbf{Data Preparation.}~~In order to acquire large-scale medical dialogues with annotated terminology, we conduct a series of preprocessing stages including outlier filtering and segment truncation on a real medical dialogue dump from MedDialog~\cite{meddg}. We then leverage the external terminology word-list for our terminology identification task, which is mainly based on distant supervision (details in the repository). Finally, we obtain a corpus containing nearly 250,000 doctor-patient conversation pairs with 3,600M tokens. We set up our experiment based on this terminology-enhanced dataset, whose statistics are shown in \autoref{tab:data_stat}. The train/val/test are split according to the ratio of 0.9/0.05/0.05. In addition, the dataset is fully shuffled to guarantee the data distribution learned by the language models can be utilised in validation and testing.

\subsection{Response Generation}
We employ BART \cite{lewis2019bart} (encoder-decoder structure) as the base model. The features of each input sequence are encoded with a self-attention mechanism, and those features are autoregressively decoded in order to predict responses, where the predicted tokens are forced to fit the gold standard responses from real doctors:
\begin{align}
    P(y_t|y_{<t},X)&=\text{softmax}(WO_t+b)\\
    O_t &= \text{Decoder}(y_{<t},F)\\
    F &= \text{Encoder}(X_\tau)\\
    \mathcal{L}_{lm} &= - \frac{1}{N}\sum_{n=1}^N \log P(Y|X)
\end{align}
where $F$ denotes the encoded features with self-attention, $O_t$ denotes the decoded feature at $t$-th position, and $W$ and $b$ are trainable parameters. $\mathcal{L}_{lm}$ denotes the cross entropy loss between predicted tokens and the reference response.

\subsection{Auxiliary Terminology-aware Training}
To better support the encoder in learning the feature distribution of terminologies, an auxiliary classification task is introduced to train the encoder to identify terminology-related tokens. This training aims to reduce the terminology classification loss $\mathcal{L}_{classifier}$ defined as follows:
\begin{align} 
    \widetilde{e_i} &= \text{classifier}(F_i)\\
    \mathcal{L}_{classifier}&=-e\log p(\widetilde{e_i})-(1-e)\log p(1-\widetilde{e_i})
\end{align}
where the encoded feature for $i$-th token $F_i$ is mapped to be a label variable $e_i$ denoting if it is domain-specific terminology (followed by $\mathrm{[TERM]}$). $\mathrm{[classifier]}$ is implemented by a fully connected network. 

\subsection{Training}
Finally, both aforementioned task objectives are combined to train the language model, and the overall loss function $\mathcal{L}_{overall}$ is minimized during model fine-tuning:
\begin{equation}
\mathcal{L}_{overall} = \mathcal{L}_{lm} + \mathcal{L}_{classifier}
\end{equation}

\section{Experiment}

\begin{table*}[ht]
\centering \small
\resizebox{0.82\linewidth}{!}{
\begin{tabular}{l|c|cccc|ccc|cccc}
\toprule
\textbf{Model} & \textbf{PPL$\downarrow$} & \textbf{B-1$\uparrow$} & \textbf{B-2$\uparrow$} & \textbf{B-3$\uparrow$} & \textbf{B-4$\uparrow$} & \textbf{R-1$\uparrow$} & \textbf{R-2$\uparrow$} & \textbf{R-L$\uparrow$} & \textbf{Dist-1$\uparrow$} & \textbf{Dist-2$\uparrow$} & \textbf{Dist-3$\uparrow$} & \textbf{Dist-4$\uparrow$} \\
\hline
    \textbf{GPT-2} & 4.7667 & 0.0725 & 0.0376 & 0.0267 & 0.0218 & 12.5431 & 4.3497 & 9.8125 & 0.0048 & 0.0245 & 0.0511 & 0.0725 \\
    \textbf{DialoGPT} & 6.2353 & 0.0599 & 0.031 & 0.0225 & 0.0187 & 9.9041 & 3.5320 & 8.0158 & 0.0036 & 0.0169 & 0.0354 & 0.0531 \\
\hline
    \textbf{T5-base} & 2.0730 & 0.1255 & 0.0585 & 0.0319 & 0.0188 & 12.7913 & 2.1351 & 9.8115 & 0.0026 & 0.0102 & 0.0183 & 0.0265 \\
    \textbf{T5-large} & 2.770 & 0.0952 & 0.0529 & 0.0388 & 0.0321 & 16.1452 & \textbf{5.8471} & 12.3637 & 0.0055 & 0.0282 & 0.0593 & 0.0883 \\
\hline
    \textbf{BART-base} & 2.7937 & 0.1205 & 0.0631 & 0.0423 & 0.0321 & 19.3620 & 5.1807 & 11.4057 & 0.0046 & 0.0374 & 0.1125 & 0.2087 \\
    \textbf{BART-large} & 3.0260 & 0.1142 & 0.0621 & 0.0420 & 0.0319 & 19.4735 & 5.3347 & 11.4324 & 0.0055 & 0.0435 & 0.1131 & 0.1933 \\ 
\hline
    \textbf{Ours (BART)} \\
    \textbf{-w terms} & 1.2367 & 0.1415 & 0.0744 & 0.0488 & 0.0359 & \textbf{21.4796} & 5.683 & 12.1532 & 0.0048 & 0.0335 & 0.0823 & 0.1403 \\
    \textbf{-w terms+AL} & \textbf{1.1419} & \textbf{0.1547} & \textbf{0.0822} & \textbf{0.0555} & \textbf{0.0421} & 20.2111 & 5.4167 & \textbf{13.0137} & \textbf{0.0071} & \textbf{0.0453} & \textbf{0.1462} & \textbf{0.2899} \\
\bottomrule
\end{tabular}
}
\caption{\label{auto evaluation}
Automatic evaluation with popular metrics used in the task of open domain dialogue generation. \textbf{AL} denotes the \textbf{A}uxiliary task for terminological representation \textbf{L}earning. The best-performing model for each metric is highlighted in bold. Note that perplexity may not be precise as different models may have different vocabulary sizes.
}
\end{table*}

\subsection{Experimental Setup}

We compare our framework with several competitive models used in prior works \cite{zeng-etal-2020-meddialog,zhou-etal-2021-generation,wang-etal-2021-fast} and related areas~\cite{tang2022ngep,huang-etal-2022-improving,tang-etal-2022-etrica}.
\begin{itemize}
[noitemsep,nolistsep,leftmargin=*]
    \item \textbf{BART} \cite{lewis2019bart}: One of the most popular language models in the text generation field.
    \item \textbf{T5} \cite{raffel2020exploring}: A widely used language model with an encoder-decoder architecture which has been successfully adapted to various generation tasks.
    \item \textbf{GPT-2} \cite{radford2019language}: A popular pre-trained language model widely adopted in tasks of dialogue generation.
    \item \textbf{DialoGPT} \cite{zhang2019dialogpt}: A dialogue generation oriented pre-trained  GPT-2 model, which gives strong performance on dialogue generation.
\end{itemize}

To comprehensively evaluate the effectiveness of our proposed framework, a range of referenced and unreferenced metrics are selected for the following experiments. Perplexity (PPL) is used to measure the language perplexity when models predict the tokens from reference responses. BLEU (B-$n$) \cite{Bleu} and ROUGE (R-$n$)\footnote{R-L refers to the longest common subsequence.} \cite{lin2004rouge}  compare the generated response with the reference by calculating their $n$-gram overlaps. We also follow \cite{li2015diversity} to measure the response quality by calculating the $n$-gram distinction as the diversity (D-$n$) of generated responses, where $n$ also denotes the $n$-gram.

\subsection{Implementation Details}

All of the pre-trained models used are based on publicly available checkpoints on Hugging Face \footnote{\url{https://huggingface.co/models}}. Models are trained for up to $10$ epochs on a Tesla A100 machine for approximately one day, and the best checkpoints are selected based on the perplexity of generated responses during validation. The batch size is set to $36$, and the learning rate is $1e^{-4}$, with the Adam optimizer selected for training.

\subsection{Experimental Results}
\label{sec:exp}
\vspace{-1mm}
As shown in Table \ref{auto evaluation}, our proposed framework (-w terms+AL) that incorporates both the terminologically enhanced dataset and auxiliary task learning outperforms all baseline models on almost all metrics (with the result on R-2 also being competitive). The substantial improvement on the metrics of BLEU and ROUGE  (some of them even experiencing about a 20$\%$ gain) demonstrates that with the incorporation of terminology features, the language model is able to generate responses more similar to gold references. Additionally, the large increase in distinction scores also indicates that the language quality has been substantially improved via the inclusion of a wider domain-specific vocabulary, resulting from the better understanding of terminological semantics. The experiments also analyse if the models benefit when the model size increases (from base to large). To our surprise, according to the metrics, the increasing model size does not result in much improvement to performance, and performance on some metrics even slightly decreases (potentially due to overfitting). We suppose that there is an upper boundary regarding model size for the vanilla language model to incorporate semantic features, after which diminishing returns are experienced. Existing language models struggle to learn domain-specific knowledge without the explicit modeling of medical concepts, and this is also how we manage to fill in the gap in terminological representations.

In the ablation study, the results further show the advances of the proposed terminological representation learning. With the simple knowledge enhancement of medical terminology (-w terms), the base language model attains substantial improvements on all referenced metrics, e.g. the language perplexity has a large decrease from about 2.79 to 1.24. It can be observed that learning feature distributions of terminology contributes a lot to the semantic understanding of medical dialogues. Furthermore, benefiting from the auxiliary training (-w terms+AL), performance on all metrics is increased by a large margin. For example, some D-n scores double (-w terms), which demonstrates the efficiency and effectiveness of incorporating features in our proposed framework.  
\begin{figure}[tb]
\centering
\includegraphics[width=0.80\columnwidth]{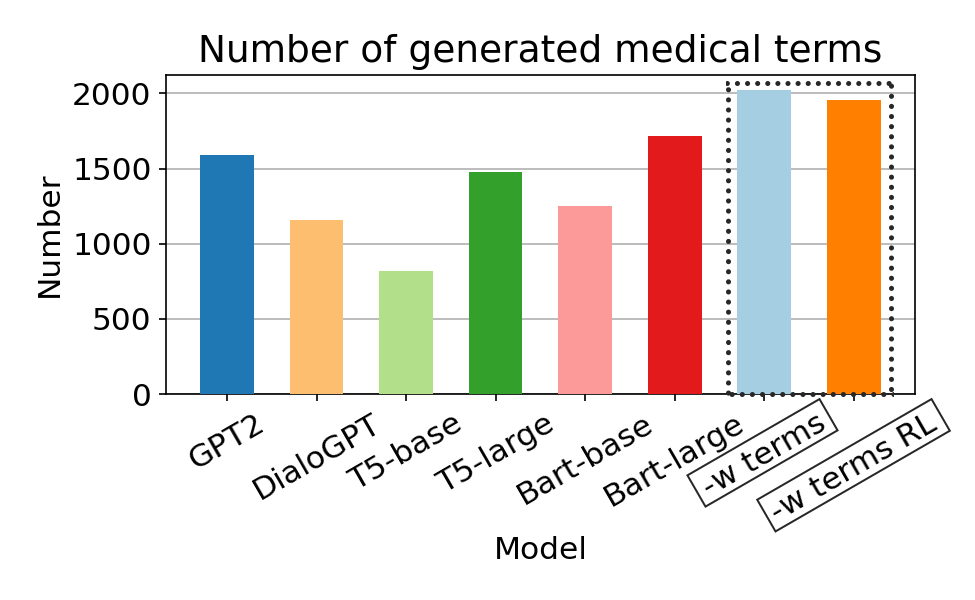}
\caption{Distinct terminologies contained in generated responses on the test dataset.}
\label{fig: Number of different terms in responses.}
\end{figure}


As shown in Fig.~\ref{fig: Number of different terms in responses.}, the terminologies contained in generated responses are also analysed to estimate the benefit of terminology-aware training. It can be observed that models that are good at incorporating terminology features tend to generate responses with a higher variety of medical terminologies. More informative responses containing additional medical knowledge are therefore beneficial for applications in telemedicine. The results also indicate the advances in incorporating domain-specific knowledge, where our proposed framework outperforms all baselines. 

\vspace{-2mm}
\section{Conclusion}
\vspace{-1mm}
In this paper, we proposed a novel framework to perform terminology-aware medical dialogue generation. We exploit the features of terminology distributions to force a language model to pay more attention to the terminology-centered parts, and generate a better response. A range of experiments demonstrate that the self-attention mechanism placed upon terminological knowledge, in addition to the auxiliary training task, can substantially improve the performance of language models on medical dialogue generation. 
We also provide the terminology annotation framework, and the corresponding large-scale dataset for further study.
\vspace{-2mm}
\section*{Acknowledgements}
\vspace{-1mm}
The authors acknowledge the support of China Scholarship Council (CSC) (File No.202006120039) and UK Research and
Innovation [grant number EP/S023062/1].

\vfill\pagebreak

\bibliographystyle{IEEEbib}
\bibliography{refs}

\end{document}